\pdfoutput=1

\documentclass[11pt]{article}

\usepackage[preprint]{acl}

\usepackage{times}
\usepackage{latexsym}

\usepackage[T1]{fontenc}

\usepackage[utf8]{inputenc}

\usepackage{microtype}

\usepackage{inconsolata}

\usepackage{graphicx}
\usepackage{amsmath,amsfonts,amssymb}
\usepackage{bbm}
\usepackage{booktabs}
\usepackage{multirow}
\usepackage{pifont}
\usepackage{tcolorbox} 
\usepackage{algpseudocode} 
\usepackage{listings} 
\usepackage{algorithm} 
\usepackage[dvipsnames]{xcolor}
\usepackage[normalem]{ulem}

\newcommand{\xmark}{\ding{55}} 
\newcommand{\cmark}{\ding{51}} 

\newcommand{\methodname}{\texttt{CoherentAD\xspace}}

\usepackage[subtle]{savetrees} 

\usepackage{xspace}
\makeatletter
\DeclareRobustCommand\onedot{\futurelet\@let@token\@onedot}
\def\@onedot{\ifx\@let@token.\else.\null\fi\xspace}

\makeatother

\usepackage[capitalize]{cleveref}
\crefname{figure}{Fig.}{Figs.}
\Crefname{figure}{Figure}{Figures}
\crefname{section}{Sec.}{Secs.}
\Crefname{section}{Section}{Sections}
\crefname{table}{Tab.}{Tabs.}
\Crefname{table}{Table}{Tables}

\title{More than a Moment: Towards Coherent Sequences of Audio Descriptions}

\author{
 \textbf{Eshika Khandelwal\textsuperscript{1}},
 \textbf{Junyu Xie\textsuperscript{2}},
 \textbf{Tengda Han\textsuperscript{2}},
 \textbf{Max Bain\textsuperscript{2}},
\\
 \textbf{Arsha Nagrani\textsuperscript{2}},
 \textbf{Andrew Zisserman\textsuperscript{2}},
 \textbf{G\"ul Varol\textsuperscript{2,3}}, 
 \textbf{Makarand Tapaswi\textsuperscript{1}}
\\
\small{
 \textsuperscript{1}CVIT, IIIT Hyderabad \hspace{5mm}
 \textsuperscript{2}VGG, University of Oxford \hspace{5mm}
 \textsuperscript{3}LIGM, École des Ponts, IP Paris, UGE, CNRS}
}

\begin{document}

\twocolumn[{%
\renewcommand\twocolumn[1][]{#1}%
\maketitle
\begin{center}
\centering
\vspace{-17mm}
\captionsetup{type=figure,skip=0pt} 
\includegraphics[width=0.97\textwidth]{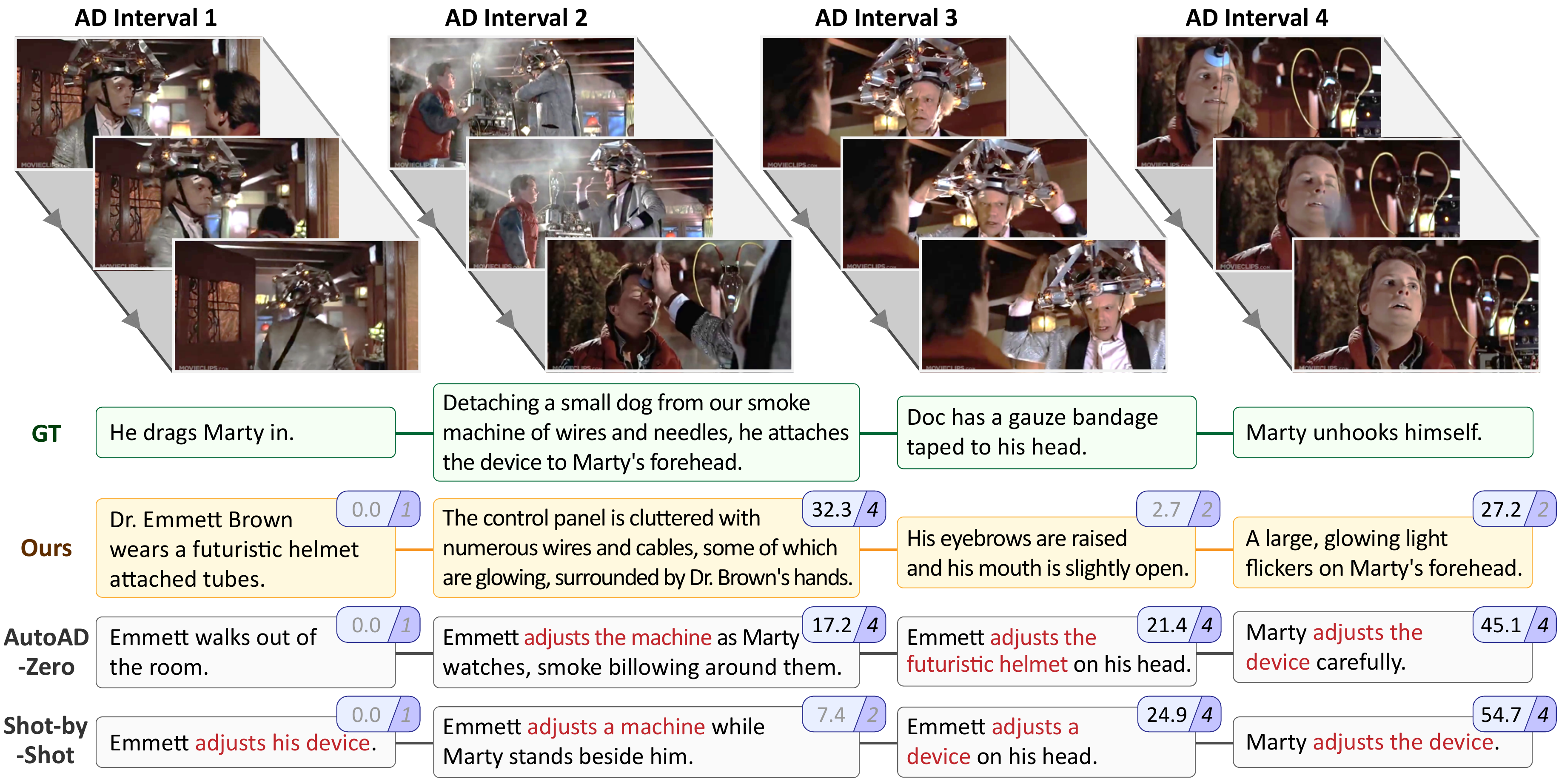}
\vspace{2mm}
\captionof{figure}{\textbf{Predicted ADs across the video} (i.e.\ a sequence of AD intervals).
The results reported by per-AD evaluation metrics are shown on the top right of each prediction (left: CIDEr; right: LLM-AD-Eval, score 1-5), with low scores indicating poor performance coloured in \textcolor{gray}{grey}.
The repetitions across predictions are highlighted in \textcolor{BrickRed}{red}, where ``adjust the device'' is repeated multiple times.
The video is sampled from the movie \textit{Back to the Future}, corresponding to $0\mathpunct{:}22-1\mathpunct{:}05$, that can be watched here: {\small\url{https://www.youtube.com/watch?v=SR5BfQ4rEqQ&t=22s}}.}
\label{fig:teaser}
\end{center}
\vspace{0.5cm}
}]

\begin{abstract}
Audio Descriptions (ADs) convey essential on-screen information, allowing visually impaired audiences to follow videos.
To be effective, ADs must form a coherent sequence that helps listeners to visualise the unfolding scene, rather than describing isolated moments.
However, most automatic methods generate each AD independently, often resulting in repetitive, incoherent descriptions.
To address this, we propose a training-free method, \methodname, that first generates multiple candidate descriptions for each AD time interval, and then performs auto-regressive selection across the sequence to form a coherent and informative narrative.
To evaluate AD sequences holistically, we introduce a sequence-level metric, StoryRecall, which measures how well the predicted ADs convey the ground truth narrative, alongside repetition metrics that capture the redundancy across consecutive AD outputs.
Our method produces coherent AD sequences with enhanced narrative understanding, outperforming prior approaches that rely on independent generations.
\end{abstract}

\section{Introduction}
\label{sec:intro}
Audio Descriptions (ADs) help the visually impaired
follow a movie or a
long video typically conveying a story.
Narrated between dialogues, ADs often describe key visual elements of the scene, with emphasis on the setting, actions, and characters.
While ADs are typically created by professionals~\cite{visual_made_verbal}, there is growing interest in automatic generation~\cite{Han23,Han23a,Han24,Xie24b,distinctad,narrad}. 

AD generation is historically treated as video captioning, i.e.~a short description is generated independently for each predefined \textit{AD interval} in the video~\cite{Rohrbach2017, soldan2022mad, Han24}.
However, ADs are a coherent sequence of descriptions that build a visual story and take the narrative forward.
As seen in \cref{fig:teaser}, this independent generation (e.g.~AutoAD-Zero~\cite{Xie24b}, Shot-by-Shot~\cite{shotbyshot}) results in repeating similar information, and failing to capture the narrative structure of the movie.

On the other hand, mainstream AD evaluations are typically conducted on a \textit{per-AD} basis, with CIDEr~\cite{Vedantam_2015_CVPR} and LLM-AD-Eval~\cite{Han24} being widely adopted. As shown in~\cref{fig:teaser}, both metrics produce highly correlated scores and often reward predictions that simply mention correct names or objects, while failing to penalise redundancy across outputs or capture the coherence of the overall narrative. Moreover, these metrics enforce matching against a single ground truth, overlooking the fact that each time interval may encompass multiple valid descriptions.

We therefore posit that both AD generation and evaluation should be performed over a longer temporal extent, i.e.\ across the {\em video} that consists of a {\em sequence} of {\em AD intervals}. 
This is motivated by the subjective nature of ADs, where information is often distributed across multiple descriptions (see \cref{fig:teaser}).

In this work, we propose a new training-free method, \methodname~that encourages the generation of diverse visual descriptions across the video.
Similar in spirit to AutoAD-Zero~\cite{Xie24b}, we first extract structured information from each trimmed clip.
By contrast, we generate multiple AD-like candidate descriptions for each clip and auto-regressively choose one that would advance the narrative while also providing new visual details.

For evaluation, we move away from conventional metrics
that compare ground truth (GT) and predicted ADs for a single interval.
Instead, we adopt:
(i)~\textit{StoryRecall} that captures whether visual details and narrative points mentioned in the GT are conveyed by the predictions; and
(ii)~\textit{Repetition} metrics that assess the redundancy of generated ADs.
We evaluate sequence-level AD generation on CMD-AD and TV-AD videos and observe qualitative and quantitative improvements in generated ADs, 
further validated through user studies.

\section{Related Work}
\label{sec:relwork}

\noindent\textbf{AD generation}~aims to produce concise, coherent narrations of the salient visual content that complement auditory signals.
Prior work falls into two categories:
(i)~\textit{End-to-end models}~\cite{Han23, Han23a, Han24, wang2024loco-mad, movieseq, uniad, distinctad, ye2025focusedad}, which are fine-tuned on domain-specific AD datasets~\cite{soldan2022mad,Han23,Xie24}; 
(ii)~\textit{Training-free frameworks}~\cite{Zhang_2024_CVPR, ye-etal-2024-mmad-multi, chu2024llmadlargelanguagemodel, Xie24b, narrad}, which adopt multi-stage setups built on pre-trained 
Vision-Language Models (VLMs) and Large Language Models (LLMs).

While early methods generate ADs independently for each time segment, recent works focus on maintaining narrative coherence and reducing redundancy across consecutive outputs.
AutoAD-I~\cite{Han23} and UniAD~\cite{uniad} adopt recursive generation processes that condition on previous AD outputs. 
AutoAD-II~\cite{Han23a} trains a localisation module to predict temporal segments for AD injection.
DistinctAD~\cite{distinctad} jointly processes adjacent AD clips and reduces redundancy via a Contextual Expectation-Maximisation Attention mechanism.
Our work explores coherent AD generation in a training-free setup.

\vspace{3pt}\noindent\textbf{AD evaluation.}
Early AD evaluation adopts captioning metrics, including n-gram overlap metrics such as CIDEr~\cite{Vedantam_2015_CVPR}, ROUGE~\cite{lin-2004-rouge}, BLEU~\cite{bleu_score}, and METEOR~\cite{banerjee-lavie-2005-meteor}, as well as semantic-oriented ones like SPICE~\cite{spice2016} and BERTScore~\cite{bert-score}.
More recent efforts have introduced AD-specific evaluations, such as retrieval-based (e.g.\ Recall@k/N~\cite{Han23}) and LLM-based assessments~\cite{Han24, Zhang_2024_CVPR}. Other metrics focus on specific aspects of AD quality, for example, CRITIC~\cite{Han24} for character accuracy and ``Action Score''~\cite{shotbyshot} for action groundedness.

Beyond single-AD evaluation, few metrics assess coherence and redundancy across consecutive ADs.
\citet{movieseq} measure n-gram repetition using R@4%
, while \citet{ye2025focusedad} introduce a redundancy-aware metric based on semantic similarity.
\citet{Zhang_2024_CVPR} propose SegEval to score short AD windows using GPT-4~\cite{openai2024gpt4technicalreport}; while \cite{adqa}~proposes a question-answering based evaluation.
In this work, we 
propose a suite of metrics for multi-AD evaluations, focusing on repetitions and overall storyline coherence.

\section{Sequence-Level AD Generation}
\label{sec:method}

\begin{figure*}
\centering
\includegraphics[width=.93\textwidth]{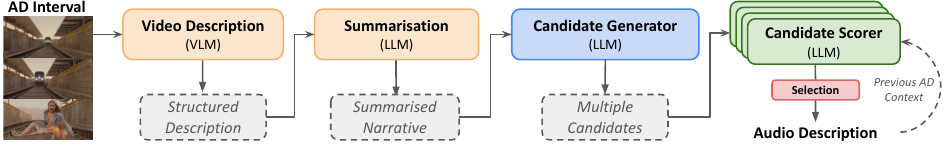}
\vspace{-0.25cm}
\caption{\textbf{Overview of our multi-stage AD generation pipeline \methodname.} For each AD interval, the VLM generates a structured description, which is then summarised. The summary is used to produce multiple candidate descriptions. Each candidate is scored by four independent LLM-based scorers that consider previous selections as context. The highest-scoring candidate is selected in an auto-regressive manner to form a coherent sequence.}
\label{fig:pipeline}
\vspace{-0.4cm}
\end{figure*}

Given a sequence of predefined video intervals, our goal is to generate the corresponding AD texts.
We introduce \methodname, a \textit{training-free} method that generates multiple candidate ADs per interval and selects a coherent, non-redundant sequence of ADs. 
The method consists of three stages (see \cref{fig:pipeline}; ablated in~\cref{subsec:app:stage_ablation}):
(i)~video interval description;
(ii)~multiple AD generation for each interval;
and
(iii)~selection of the optimal AD sequence.

\subsection{Video interval to summarised narrative}
\label{subsec:extract_filter}
We describe the visual frames 
within each AD interval by first extracting structured textual descriptions of the visual content with a VLM, and then by summarising them with an LLM.
Specifically, we
first
prompt a VLM using a three-part instruction (see~\cref{subsec:app:stage1prompts}) to extract all visually relevant details in a 
structured format. 
Since later stages rely solely on text, any missed detail is unrecoverable, making it essential to capture a complete description at this stage. 
This description also supports diverse candidate generation later in the pipeline.
Second,
we use an LLM to rephrase 
the exhaustive structured visual description
into a concise,
yet complete paragraph $P$ that preserves all relevant content.
This serves as the basis for the next stage.

\subsection{Multiple AD candidates per interval}
\label{subsec:candidate_generation}
Given the paragraph $P$,
we prompt an LLM to generate up to $m$ diverse candidate ADs 
that are (i)~independent, (ii)~concise, (iii)~group related visual events, and (iv)~collectively cover the full content from the previous stage.
Each candidate conveys a complete visual moment by grouping related elements (such as actions, objects and context) into a compact description.
This makes every candidate
suitable for inclusion in the final sequence on its own.
Not forcing a fixed number, but allowing less than $m$ candidates avoids redundancy across candidates, while preventing fragmentation.

\subsection{Coherent sequence selection}
\label{subsec:sequence_selection}
Finally, to form a sequence of AD predictions over an input video, we perform auto-regressive selection across intervals, conditioning on previously selected ADs.
This selection process is guided by four scoring criteria, with each score assigned independently by a separate LLM scorer to ensure focused and unbiased judgment for each aspect.
(i)~\textit{Adherence to AD guidelines}~\citep{visual_made_verbal}
checks if the description 
is
strictly grounded in what is visually perceivable (e.g.\ no inferences, speculation, or camera references) and
focuses only on what a visually impaired viewer cannot directly access.
(ii)~\textit{Redundancy}
penalises repeated content from prior descriptions. 
(iii)~\textit{Story advancement} prioritises candidates that introduce new observable actions, interactions, or scene changes that move the 
narrative forward.
(iv)~\textit{Counts of visual elements}
tallies the number of unique participants, actions, and salient visual details explicitly mentioned in the description, rewarding candidates that convey more information.

The final score for each candidate is computed as a weighted average of these four criteria, and the highest-scoring one is selected auto-regressively, conditioned on the previous $r$ selected descriptions.

\begin{table*}
\centering
\setlength\tabcolsep{7pt}
\resizebox{\linewidth}{!}{
\begin{tabular}{lc ccc ccc}
\toprule
\multirow{2}{*}{Method} & \multirow{2}{*}{Train} & \multicolumn{3}{c}{\textbf{CMD-AD}} & \multicolumn{3}{c}{\textbf{TV-AD}} \\
\cmidrule(lr){3-5} \cmidrule(lr){6-8}
 & Free & SR $\uparrow$ & Exact Repeat \% $\downarrow$ & Partial Repeat \% $\downarrow$ 
 & SR $\uparrow$ & Exact Repeat \% $\downarrow$ & Partial Repeat \% $\downarrow$ \\

\midrule
Ground Truth  & - & - & (\textcolor{white}{0.0}0, \textcolor{white}{0.0}0, \textcolor{white}{0.0}0) & (\textcolor{white}{0}3.71, \textcolor{white}{0}4.42, \textcolor{white}{0}3.97) & - & (\textcolor{white}{0.0}0, \textcolor{white}{0.0}0, \textcolor{white}{0.0}0) & (\textcolor{white}{0}4.56, \textcolor{white}{0}4.38, \textcolor{white}{0}3.70)\\
\midrule
AutoAD-III~\cite{Han24} & \xmark & 2.11 & (4.51, 2.81, 1.99) & (16.01, 11.90, 10.44)   & - & - & - \\
UniAD~\cite{uniad}   & \xmark & 2.13 & (4.15, 2.44, 2.10) & (16.21, 12.27, 11.57)  & - & - & - \\
\midrule
AutoAD-Zero$\dagger$~\cite{Xie24b}  & \cmark  & 2.27   & (0.19, 0.32, 0.30)  & (\textcolor{white}{0}7.87, \textcolor{white}{0}8.13, \textcolor{white}{0}8.13)   & 1.69 & 
(1.15, 1.23, 0.32)
& (21.78, 17.47, 14.83)\\
AutoAD-Zero & \cmark  & 2.27   & (0.55, 0.31, 0.21) & (13.26, 10.62, \textcolor{white}{0}9.59)  & 1.79 & (0.89, 0.18, 0.16) & (19.23, 14.90, 13.74) \\
Shot-by-Shot~\cite{shotbyshot}  & \cmark  & 2.43  & (0.42, 0.15, 0.12)  & (19.06, 15.18, 13.56)  & \textbf{1.84} & (0.73, 0.35, 0.16) & (21.19, 15.89, 13.46) \\
\midrule
\methodname~(Ours) & \cmark & \textbf{2.63}  & (\textcolor{white}{0.0}\textbf{0}, \textcolor{white}{0.0}\textbf{0}, \textcolor{white}{0.0}\textbf{0})  & (\textcolor{white}{0}\textbf{5.21}, \textcolor{white}{0}\textbf{4.45}, \textcolor{white}{0}\textbf{4.18})   & 1.83 & (\textcolor{white}{0.0}\textbf{0}, \textcolor{white}{0.0}\textbf{0}, \textcolor{white}{0.0}\textbf{0})  & (\textcolor{white}{0}\textbf{6.16}, \textcolor{white}{0}\textbf{5.36}, \textcolor{white}{0}\textbf{4.26}) \\
\quad w/o multiple candidates & \cmark & 2.49 & (0.04, \textcolor{white}{0.0}\textbf{0}, 0.02) & (\textcolor{white}{0}8.17, \textcolor{white}{0}6.78, \textcolor{white}{0}6.19)  & - & - & - \\
\bottomrule
\end{tabular}
}
\vspace{-0.3cm}
\caption{\textbf{Quantitative comparison on CMD-AD and TV-AD.}
The first row indicates the inherent level of repetition in ground-truth descriptions, serving as a lower bound for repetition scores. $\dagger$ denotes the original AutoAD-Zero adopting the VideoLLaMA-7B VLM backbone in the first stage, while all other training-free methods (including \textit{Ours}) employ Qwen2-VL-7B. 
The repetitions are reported against the three offsets.
SR denotes StoryRecall.
}
\vspace{-0.4cm}
\label{tab:sota}
\end{table*}

\section{Sequence-Level AD Evaluation}
\label{sec:metrics}
Traditional AD evaluation independently compares each description to a single ground truth (GT)~\citep{adqa}.
This overlooks the fact that ADs are intended to form a coherent sequence 
that, among other goals,
(i)~conveys the story, 
and (ii)~remains non-redundant.
To address this, we propose two metrics.
\textit{StoryRecall} evaluates whether the generated sequence captures the same visual story as the reference GT sequence 
(\cref{subsec:storyrecall}),
and \textit{repetition} metrics measure redundancy.
(\cref{subsec:repetition}).

\subsection{StoryRecall}
\label{subsec:storyrecall}
To evaluate whether the predicted AD sequence captures the key events of the video, we assess how well they recover the storyline conveyed in the GT sequence.
Although individual GT descriptions may not align one-to-one with the predicted ADs~\cite{adqa},
the full sequence collectively captures the core visual events, making it a reliable reference.

We concatenate GT and predicted ADs for each interval and compare resulting AD sequences using an LLM.
A score from $1$ to $5$ reflects how much of the GT's visual content is conveyed.
For example, a score of $5$ indicates that the predicted sequence captures nearly all key actions, events, and visual details described in the GT.
Note, during the comparison, paraphrasing and reordering are allowed, provided the core visual narrative remains intact.
Extra information in the predicted sequence is not penalised.
\subsection{Repetition metrics}
\label{subsec:repetition}
We also propose to explicitly monitor repetition in AD evaluation, as it reduces the effectiveness of an AD sequence by wasting valuable narration time and limiting the inclusion of new information. 
Specifically, we consider two simple repetition measures.

First, we compute the number of \textit{exact repetitions}. For each description, we compare it to the next three consecutive descriptions and check for exact string matches. We then report the proportion of ADs with exact matches across the entire dataset, yielding three percentages---one for each offset.

Second, we capture \textit{partial repetitions} through lexical overlap, 
by computing the intersection over union between descriptions.
Specifically,
we extract a set of tokens from each description
by lowercasing the text, removing punctuation and English stopwords, and applying tokenisation using NLTK~\cite{nltk}.
As with exact repeats, we report three scores for the three offsets.

\section{Experiments}
\label{sec:experiments}

\noindent\textbf{Datasets and details.}
We evaluate on
(i)~CMD-AD~\citep{Han24}, with 7,316 ADs for 591 videos of 98 movies, and
(ii)~TV-AD~\citep{Xie24b}, with 2,983 ADs spanning 100 episodes across two TV series.
We use Qwen2-VL-7B~\citep{Qwen2VL} as the VLM and LLaMA3.1-Instruct-8B~\citep{grattafiori2024llama3herdmodels} as the LLM.
We generate up to $m {=} 5$ candidates per interval and condition scoring on the $r{=}3$ previously selected candidates.
Additional details and ablations in
\cref{sec:app:implementation,sec:app:experiments}.

\vspace{3pt}\noindent\textbf{Quantitative results.}
\cref{tab:sota} reports the performance on CMD-AD and TV-AD for both fine-tuned (top) and training-free (bottom) methods, where all prior works fall short on our sequence-level metrics.
For instance, Shot-by-Shot (SbS) produces higher
partial repetitions ($19.06\%$)
between consecutive predictions
than the ground truth ($3.71\%$).
In contrast, \methodname{} achieves repetition scores that closely match GT ($5.21\%$), with zero exact repeats.
Our method prioritises sequence-level coherence by design, outperforming previous state-of-the-art on \textit{StoryRecall} ($2.63$ vs $2.43$) on CMD-AD, while achieving comparable performance for TV-AD.

Notably, the ablation without multiple candidate generation
achieves significantly lower repetition vs SbS ($8.17\%$ vs $19.06\%$).
The latter uses concise VLM outputs highlighting the most prominent character, action, or interaction,
often leading to redundant phrasing.
Our setup aggregates wider VLM outputs, resulting in diverse and informative descriptions.
Further, shifting to our multi-candidate setup increases StoryRecall ($2.49$ to $2.63$) and decreases repetition ($8.17\%$ to $5.21\%$).

Finally, we also see good results on the ADQA benchmark~\citep{adqa} in~\cref{sec:app:adqa}.

\vspace{3pt}\noindent\textbf{Qualitative results}
are shown in \cref{fig:teaser} and \cref{sec:app:qualitative}.
Repetitions are clearly observed in baseline predictions,
while \methodname{} introduces distinct, visibly grounded details at each interval forming a coherent and non-redundant narrative.

\section{Conclusion}
\label{sec:conclusion}

We highlighted the limitations of current methods, producing repetitions and incoherence in sequential ADs.
\methodname{}, our training-free approach attempted to address this through multiple coherence criteria, and posing the problem as sequence search among multiple candidates per AD interval.

\vspace{3pt}\noindent\textbf{Acknowledgements.}
This project was funded in part by the ANR project CorVis ANR-21-CE23-0003-01 and a research gift from Google.
It was also supported by an SERB SRG/2023/002544 grant.
The authors also thank Divy Kala for evaluating generated ADs on ADQA.

\section*{Limitations}
\label{sec:limitations}

Similar to most prior work, our method relies on reference-provided temporal intervals and does not address the problem of AD localisation, i.e.\ predicting \textit{when} an AD should be placed. 
These intervals are assumed as input to the pipeline.

Our approach also does not incorporate neighbouring context during generation or selection.
However, ADs are not required to match the exact interval boundaries and can refer to nearby events.
Leveraging neighbouring context could help fill in gaps and improve coherence.

Finally, we select the best candidate per interval without post-processing. Editing or merging candidates could further enhance sequence-level fluency.

\bibliography{references,vgg_local}

\begin{thebibliography}{32}
\providecommand{\natexlab}[1]{#1}

\bibitem[{Anderson et~al.(2016)Anderson, Fernando, Johnson, and Gould}]{spice2016}
Peter Anderson, Basura Fernando, Mark Johnson, and Stephen Gould. 2016.
\newblock Spice: Semantic propositional image caption evaluation.
\newblock In \emph{ECCV}.

\bibitem[{Banerjee and Lavie(2005)}]{banerjee-lavie-2005-meteor}
Satanjeev Banerjee and Alon Lavie. 2005.
\newblock {METEOR}: An automatic metric for {MT} evaluation with improved correlation with human judgments.
\newblock In \emph{Proceedings of the {ACL} Workshop on Intrinsic and Extrinsic Evaluation Measures for Machine Translation and/or Summarization}.

\bibitem[{Bird et~al.(2009)Bird, Loper, and Klein}]{nltk}
Steven Bird, Edward Loper, and Ewan Klein. 2009.
\newblock \emph{Natural Language Processing with Python}.
\newblock O'Reilly Media.

\bibitem[{Chu et~al.(2024)Chu, Wang, and Abrantes}]{chu2024llmadlargelanguagemodel}
Peng Chu, Jiang Wang, and Andre Abrantes. 2024.
\newblock {LLM-AD}: Large language model based audio description system.
\newblock \emph{arXiv preprint arXiv:2405.00983}.

\bibitem[{{Cursor}()}]{cursor}
{Cursor}.
\newblock Cursor: Ai-powered code editor.
\newblock \url{https://www.cursor.sh}.

\bibitem[{Fang et~al.(2025)Fang, Wu, Wu, Song, and Chan}]{distinctad}
Bo~Fang, Wenhao Wu, Qiangqiang Wu, Yuxin Song, and Antoni~B. Chan. 2025.
\newblock {DistinctAD}: Distinctive audio description generation in contexts.
\newblock In \emph{CVPR}.

\bibitem[{Han et~al.(2023{\natexlab{a}})Han, Bain, Nagrani, Varol, Xie, and Zisserman}]{Han23a}
Tengda Han, Max Bain, Arsha Nagrani, G{\"u}l Varol, Weidi Xie, and Andrew Zisserman. 2023{\natexlab{a}}.
\newblock {AutoAD II}: The sequel – who, when, and what in movie audio description.
\newblock In \emph{ICCV}.

\bibitem[{Han et~al.(2023{\natexlab{b}})Han, Bain, Nagrani, Varol, Xie, and Zisserman}]{Han23}
Tengda Han, Max Bain, Arsha Nagrani, G{\"u}l Varol, Weidi Xie, and Andrew Zisserman. 2023{\natexlab{b}}.
\newblock Autoad: Movie description in context.
\newblock In \emph{CVPR}.

\bibitem[{Han et~al.(2024)Han, Bain, Nagrani, Varol, Xie, and Zisserman}]{Han24}
Tengda Han, Max Bain, Arsha Nagrani, G{\"u}l Varol, Weidi Xie, and Andrew Zisserman. 2024.
\newblock {AutoAD III}: The prequel -- back to the pixels.
\newblock In \emph{CVPR}.

\bibitem[{Kala et~al.(2025)Kala, Khandelwal, and Tapaswi}]{adqa}
Divy Kala, Eshika Khandelwal, and Makarand Tapaswi. 2025.
\newblock {What You See is What You Ask}: Evaluating audio descriptions.
\newblock In \emph{EMNLP}.

\bibitem[{Lin(2004)}]{lin-2004-rouge}
Chin-Yew Lin. 2004.
\newblock {ROUGE}: A package for automatic evaluation of summaries.
\newblock In \emph{Text Summarization Branches Out}.

\bibitem[{Lin et~al.(2024)Lin, Zhang, Gao, Xia, Chen, Gao, Xie, Xiao, and Shou}]{movieseq}
Kevin~Qinghong Lin, Pengchuan Zhang, Difei Gao, Xide Xia, Joya Chen, Ziteng Gao, Jinheng Xie, Xuhong Xiao, and Mike~Zheng Shou. 2024.
\newblock Learning video context as interleaved multimodal sequences.
\newblock In \emph{ECCV}.

\bibitem[{Meta(2024)}]{grattafiori2024llama3herdmodels}
Meta. 2024.
\newblock The {L}lama 3 herd of models.
\newblock \emph{arXiv preprint arXiv: 2407.21783}.

\bibitem[{OpenAI()}]{chatgpt}
OpenAI.
\newblock Chatgpt: Language model for dialogue.
\newblock \url{https://openai.com/chatgpt}.

\bibitem[{OpenAI(2024)}]{openai2024gpt4technicalreport}
OpenAI. 2024.
\newblock Gpt-4 technical report.
\newblock \emph{arXiv preprint arXiv: 2303.08774}.

\bibitem[{Papineni et~al.(2002)Papineni, Roukos, Ward, and Zhu}]{bleu_score}
Kishore Papineni, Salim Roukos, Todd Ward, and Wei-Jing Zhu. 2002.
\newblock Bleu: a method for automatic evaluation of machine translation.
\newblock In \emph{{ACL}}.

\bibitem[{Park et~al.(2025)Park, Ye, Lee, Ka, and Han}]{narrad}
Jaehyeong Park, Juncheol Ye, Seungkook Lee, Hyun~W. Ka, and Dongsu Han. 2025.
\newblock {NarrAD}: Automatic generation of audio descriptions for movies with rich narrative context.
\newblock In \emph{WACV}.

\bibitem[{Rohrbach et~al.(2017)Rohrbach, Torabi, Rohrbach, Tandon, Pal, Larochelle, Courville, and Schiele}]{Rohrbach2017}
Anna Rohrbach, Atousa Torabi, Marcus Rohrbach, Niket Tandon, Christopher Pal, Hugo Larochelle, Aaron Courville, and Bernt Schiele. 2017.
\newblock Movie description.
\newblock \emph{IJCV}.

\bibitem[{Snyder(2014)}]{visual_made_verbal}
Joel Snyder. 2014.
\newblock \emph{{The Visual Made Verbal: A Comprehensive Training Manual and Guide to the History and Applications of Audio Description}}.
\newblock Dog Ear Publishing, LLC.

\bibitem[{Soldan et~al.(2022)Soldan, Pardo, Alc{\'a}zar, Caba, Zhao, Giancola, and Ghanem}]{soldan2022mad}
Mattia Soldan, Alejandro Pardo, Juan~Le{\'o}n Alc{\'a}zar, Fabian Caba, Chen Zhao, Silvio Giancola, and Bernard Ghanem. 2022.
\newblock {MAD}: A scalable dataset for language grounding in videos from movie audio descriptions.
\newblock In \emph{CVPR}.

\bibitem[{Vedantam et~al.(2015)Vedantam, Lawrence~Zitnick, and Parikh}]{Vedantam_2015_CVPR}
Ramakrishna Vedantam, C.~Lawrence~Zitnick, and Devi Parikh. 2015.
\newblock {CIDEr}: Consensus-based image description evaluation.
\newblock In \emph{CVPR}.

\bibitem[{Wang et~al.(2025)Wang, Tong, Zheng, Shen, and Wang}]{uniad}
Hanlin Wang, Zhan Tong, Kecheng Zheng, Yujun Shen, and Limin Wang. 2025.
\newblock Contextual {AD} narration with interleaved multimodal sequence.
\newblock In \emph{CVPR}.

\bibitem[{Wang et~al.(2024{\natexlab{a}})Wang, Liu, and Wu}]{wang2024loco-mad}
Jiayi Wang, Zihao Liu, and Xiaoyu Wu. 2024{\natexlab{a}}.
\newblock {LoCo-MAD}: Long-range context-enhanced model towards plot-centric movie audio description.
\newblock In \emph{ACCV}.

\bibitem[{Wang et~al.(2024{\natexlab{b}})Wang, Bai, Tan, Wang, Fan, Bai, Chen, Liu, Wang, Ge, Fan, Dang, Du, Ren, Men, Liu, Zhou, Zhou, and Lin}]{Qwen2VL}
Peng Wang, Shuai Bai, Sinan Tan, Shijie Wang, Zhihao Fan, Jinze Bai, Keqin Chen, Xuejing Liu, Jialin Wang, Wenbin Ge, Yang Fan, Kai Dang, Mengfei Du, Xuancheng Ren, Rui Men, Dayiheng Liu, Chang Zhou, Jingren Zhou, and Junyang Lin. 2024{\natexlab{b}}.
\newblock Qwen2-vl: Enhancing vision-language model's perception of the world at any resolution.
\newblock \emph{arXiv preprint arXiv:2409.12191}.

\bibitem[{Xie et~al.(2025)Xie, Han, Bain, Nagrani, Khandelwal, Varol, Xie, and Zisserman}]{shotbyshot}
Junyu Xie, Tengda Han, Max Bain, Arsha Nagrani, Eshika Khandelwal, G\"ul Varol, Weidi Xie, and Andrew Zisserman. 2025.
\newblock Shot-by-shot: Film-grammar-aware training-free audio description generation.
\newblock In \emph{ICCV}.

\bibitem[{Xie et~al.(2024{\natexlab{a}})Xie, Han, Bain, Nagrani, Varol, Xie, and Zisserman}]{Xie24b}
Junyu Xie, Tengda Han, Max Bain, Arsha Nagrani, G{\"u}l Varol, Weidi Xie, and Andrew Zisserman. 2024{\natexlab{a}}.
\newblock Autoad-zero: A training-free framework for zero-shot audio description.
\newblock In \emph{ACCV}.

\bibitem[{Xie et~al.(2024{\natexlab{b}})Xie, Xie, and Zisserman}]{Xie24}
Junyu Xie, Weidi Xie, and Andrew Zisserman. 2024{\natexlab{b}}.
\newblock Appearance-based refinement for object-centric motion segmentation.
\newblock In \emph{ECCV}.

\bibitem[{Xie et~al.(2024{\natexlab{c}})Xie, Yang, Xie, and Zisserman}]{Xie24a}
Junyu Xie, Charig Yang, Weidi Xie, and Andrew Zisserman. 2024{\natexlab{c}}.
\newblock Moving object segmentation: All you need is sam (and flow).
\newblock In \emph{ACCV}.

\bibitem[{Ye et~al.(2024)Ye, Chen, Li, Xin, Li, Zhou, and Bu}]{ye-etal-2024-mmad-multi}
Xiaojun Ye, Junhao Chen, Xiang Li, Haidong Xin, Chao Li, Sheng Zhou, and Jiajun Bu. 2024.
\newblock {MMAD}: Multi-modal movie audio description.
\newblock In \emph{LREC-COLING}.

\bibitem[{Ye et~al.(2025)Ye, Wang, Song, Zhou, Li, and Bu}]{ye2025focusedad}
Xiaojun Ye, Chun Wang, Yiren Song, Sheng Zhou, Liangcheng Li, and Jiajun Bu. 2025.
\newblock {FocusedAD}: Character-centric movie audio description.
\newblock \emph{arXiv preprint arXiv:2504.12157}.

\bibitem[{Zhang et~al.(2024)Zhang, Lin, Yang, Wang, Li, Lin, Liu, and Wang}]{Zhang_2024_CVPR}
Chaoyi Zhang, Kevin Lin, Zhengyuan Yang, Jianfeng Wang, Linjie Li, Chung-Ching Lin, Zicheng Liu, and Lijuan Wang. 2024.
\newblock {MM-Narrator}: Narrating long-form videos with multimodal in-context learning.
\newblock In \emph{CVPR}.

\bibitem[{Zhang* et~al.(2020)Zhang*, Kishore*, Wu*, Weinberger, and Artzi}]{bert-score}
Tianyi Zhang*, Varsha Kishore*, Felix Wu*, Kilian~Q. Weinberger, and Yoav Artzi. 2020.
\newblock Bertscore: Evaluating text generation with bert.
\newblock In \emph{ICLR}.

\end{thebibliography}

\clearpage
{\noindent \large \bf {Appendix}}\\
\appendix

\renewcommand{\thefigure}{A.\arabic{figure}}
\setcounter{figure}{0} 
\renewcommand{\thetable}{A.\arabic{table}}
\setcounter{table}{0} 

\cref{sec:app:implementation} provides
additional implementation details of our multi-stage pipeline,
\cref{sec:app:experiments} presents various additional ablations and experiments, and
\cref{sec:app:adqa} evaluates \methodname{} on the new AD evaluation benchmark, ADQA, showing promising results,
and
\cref{sec:app:qualitative} shows qualitative examples.

\section{Implementation Details}
\label{sec:app:implementation}

\subsection{Video interval to summarised narrative}
\label{subsec:app:stage1prompts}

We complement \cref{subsec:extract_filter} with additional details.

\vspace{2pt} \noindent \textbf{Extracting structured visual descriptions.}
We uniformly sample $16$ frames per AD interval, following standard practice in~\citet{Xie24b,shotbyshot}.
For character recognition, we adopt the method of~\citet{Xie24b}, which overlays coloured circles around detected faces and uses them to provide character names as text prompts.

We found that when shown a sequence of sampled frames, the VLM often defaults to static, high-level descriptions (e.g.\ ``a man is standing'') rather than capturing actions and changes unfolding over time. 
To address this, we prompt the VLM with a structured instruction that guides the model to extract all visually relevant details across three aspects---key actions, interactions, and environmental changes, respectively.
The exact prompt is provided in \cref{fig:app:stage1_vlm_prompt}, which consists of:

\begin{itemize}
	\setlength{\parskip}{0pt}
	\setlength{\itemsep}{0.2pt}
    \item \textit{Storyboard Description} - a step-by-step narration of events in order, treating the frames like a storyboard: a sequence of images that captures the key moments of a scene;
    \item \textit{Character and Object Breakdown} - a list of all visible characters and objects, along with their observable actions (both clear and subtle), interactions, and any environmental changes;
    \item \textit{Overall Summary} - a brief description of the primary event.
\end{itemize}

\vspace{2pt} \noindent \textbf{Narrative summarisation.}
We instruct the LLM to retain all meaningful visual content while rephrasing the dense structured description into a concise paragraph (using~\cref{fig:app:stage1_llm_prompt}). It explicitly discourages inference, dialogue, and speculation, resulting in a grounded description suitable for the candidate generation stage. 
To ensure adherence to the guidelines, the model drafts and iteratively refines the paragraph until it satisfies all specified constraints.
The resulting output is compact and fluent, serving as the basis for the next stage.

Overall, this two-step design allows us to first extract all relevant visual details, and then organise them into a concise, event-level narrative suitable for Audio Description.

\vspace{-0.1cm}
\subsection{Multiple candidate generation}
\vspace{-0.05cm}
As explained in \cref{subsec:candidate_generation},
we instruct the LLM to generate up to $m$ candidate diverse ADs, each having a word limit of ${l}_\text{max}$.
Here, ${l}_\text{max}$  adapted from~\citet{shotbyshot}, treated as target length rather than an upper bound, prompting each candidate to be as informative as possible.
The prompt~(\cref{fig:app:stage2}) encourages grouping related observations, resulting in each candidate conveying a complete and cohesive visual event.

\vspace{-0.1cm}
\subsection{Criterion weights for sequence selection}
\vspace{-0.05cm}
\label{subsec:app:scoring}
In~\cref{subsec:sequence_selection}, all scores are normalised to the range $[0, 1]$. We then apply the following weights: $0.40$ for \textit{Adherence to AD guidelines}, $0.25$ for \textit{Redundancy}, $0.40$ for \textit{Story Advancement}, and $0.29$ for \textit{Counts}. The ``\textit{Counts}'' score is computed by combining sub-scores for Participants ($0.13$), Actions ($0.11$), and Salient Details ($0.05$). Prompts for each criterion can be found in 
\cref{fig:app:ad_like,fig:app:redundancy,fig:app:story,fig:app:count}.
The weights were chosen to reflect the intended focus of the selection:
the highest weight is given to whether the AD follows guidelines and advances the story, followed by redundancy, and the least to the number of participants, actions, and other details.

We experimented with ten additional configurations on CMD-AD by randomly varying each weight between $0.02-0.05$ while preserving the relative ordering.
The results remain consistent, with StoryRecall at $2.60\pm0.02$ and partial repeats at ($5.27\pm0.02$, $4.41\pm0.02$, $4.14\pm0.02$), indicating that small shifts in weights do not meaningfully affect performance.

\begin{table*}[ht]
\centering
\resizebox{\linewidth}{!}{
\begin{tabular}{lccccccc}
\toprule
\multicolumn{1}{c}{Method} & 
\multicolumn{1}{c}{VLM:} & 
\multicolumn{1}{c}{LLM 1:} & 
\multicolumn{1}{c}{LLM 2:} & 
\multicolumn{1}{c}{LLM 3:} & 
\multicolumn{1}{c}{Story} & 
\multicolumn{1}{c}{Exact} & 
\multicolumn{1}{c}{Partial} \\

\multicolumn{1}{c}{Name} &
\multicolumn{1}{c}{Video Desc.} &
\multicolumn{1}{c}{Summ.} &
\multicolumn{1}{c}{Candidate Gen.} &
\multicolumn{1}{c}{Scorer} &
\multicolumn{1}{c}{Recall $\uparrow$} &
\multicolumn{1}{c}{Repeat\% $\downarrow$} &
\multicolumn{1}{c}{Repeat\% $\downarrow$} \\
\midrule
Coherent AD & \checkmark & \checkmark & \checkmark & \checkmark & 2.63 & 
(\textcolor{white}{0.0}0,
\textcolor{white}{0.0}0,
\textcolor{white}{0.0}0)
& 
(\textcolor{white}{0}5.21, \textcolor{white}{0}4.45, \textcolor{white}{0}4.18) \\
& \checkmark & -         & \checkmark & \checkmark & 2.58 & (\textcolor{white}{0.0}0,
\textcolor{white}{0.0}0,
\textcolor{white}{0.0}0) & (\textcolor{white}{0}6.23, \textcolor{white}{0}5.32, \textcolor{white}{0}4.88) \\
& \checkmark & \checkmark & \checkmark & -         & 2.49 & (0.04, \textcolor{white}{0.0}0, 0.02) & (\textcolor{white}{0}8.17, \textcolor{white}{0}6.78, \textcolor{white}{0}6.19) \\
VLM only    & \checkmark & -         & -         & -         & 2.38 & (1.37, 0.62, 0.41) & (15.31, 12.20, 10.92) \\
\bottomrule
\end{tabular}
}
\caption{\textbf{Ablating pipeline stages.} The full multi-stage pipeline, comprising video description (VLM), summarization (LLM 1), candidate generation (LLM 2), and scoring (LLM 3), achieves the best overall performance with highest story recall and lowest repetitions.}
\label{tab:app:stage_ablation}
\end{table*}

\subsection{Model sizes and hardware setup}
\vspace{-0.05cm}
\label{subsec:app:hardware}
Our pipeline uses open-source models with 7B and 8B parameters: Qwen2-VL-\textit{7B}~\citep{Qwen2VL} for extracting structured visual descriptions, and LLaMA-3.1-Instruct-\textit{8B}~\citep{grattafiori2024llama3herdmodels} for candidate generation and scoring. 
All experiments are run on NVIDIA A6000 GPUs, with a single run per setting; all reported values reflect these single runs.

\begin{table}
\centering
\small
\begin{tabular}{lr}
\toprule
Step & Seconds \\
\midrule
Structured descriptions & 14.88 \\
Summarisation & 1.46 \\
Multiple candidates & 0.66 \\
Calculating Counts & 0.87 \\
AD guidelines compliance & 1.57 \\
Story + Redundancy + recursively selecting & 2.60 \\
\midrule
\textbf{Total Time per AD} & \textbf{22.05} \\
\bottomrule
\end{tabular}
\caption{\textbf{Processing time per AD.} Breakdown of the average time (in seconds) taken by each stage in the pipeline, totalling \textit{22.05} seconds per description.}
\label{tab:app:time}
\end{table}

\subsection{Computational cost}

On average, processing a single AD interval takes approximately 22 seconds, with around 15 seconds spent on the VLM step (see~\cref{tab:app:time}, for a detailed breakdown). 
Note, this process can also be parallelised across video segments. 

We use a lightweight LLaMA3-8B model for efficient inference. 
Additionally, since AD generation is an offline process and not intended for real-time use, the computational costs are incurred only once.
Overall, the time and cost remains practical relative to typical movie production timelines and budgets.

\vspace{-0.1cm}
\subsection{Use of AI Assistants}
\vspace{-0.05cm}
We use~\citet{cursor} for code autocompletion, and~\citet{chatgpt} to test zero-shot prompts during development.

\section{Additional Experiments}
\vspace{-0.05cm}
\label{sec:app:experiments}

\begin{table*}
\centering
\small
\begin{tabular}{lc c cc cccc}
\toprule
\multirow{2}{*}{Method} & \multirow{2}{*}{Train} & \multicolumn{2}{c}{Old Metrics} & \multicolumn{2}{c}{\color{RoyalPurple}\textbf{Vis App}} & \multicolumn{2}{c}{\color{ForestGreen}\textbf{Narr Und}} \\
& & C & LLMe & CC & Ratio & CC & Ratio \\
\midrule
Dialog only   & -      & -    & -    & 10.0 & 33.1 & 58.9 & 81.0 \\
AutoAD-III    & \cmark & \textbf{25.0} & 2.01 & 14.9 & 49.3 & \textbf{\textit{63.2}} & \textbf{\textit{86.9}} \\
UniAD*        & \cmark & 21.8 & \textbf{2.92} & 14.3 & 47.4 & 63.0 & 86.6 \\
AutoAD-Zero   & \xmark & 17.7 & 1.96 & 13.4 & 44.3 & 62.9 & 86.5 \\
Q2VL          & \xmark & -    & -    & \textbf{17.2} & \textbf{57.0} & 51.2 & 70.4 \\
\methodname (Ours)  & \xmark & 13.2 & 2.17 & \textbf{\textit{15.2}} & \textbf{\textit{50.3}} & \textbf{64.0} & \textbf{88.0} \\
\midrule
AV$_1$        & -      & -    & -    & -    & -    & 72.7 & 100  \\
AV$_2$ (17)   & -      & -    & -    & 30.2 & 100  & 75.0 & 103 \\ 
\bottomrule
\end{tabular}
\caption{Evaluation of \methodname~on ADQA.
Acronyms are as follows:
{\color{RoyalPurple}Vis App: Visual Appreciation},
{\color{ForestGreen}Narr Und: Narrative Understanding}.
The old metrics that compare GT and predicted ADs one-to-one are
C: CIDEr and LLMe: LLM-AD-eval~\cite{Han24}.
The new metrics proposed in ADQA are
CC: correct answer using context, and
Ratio: Accuracy ratio.}
\label{tab:app:adqa}
\end{table*}

\subsection{Ablation of pipeline stages}
\label{subsec:app:stage_ablation}
While using a single VLM or combining all stages may seem simpler, we found it significantly reduced output quality. Each stage (event extraction, candidate generation, and scoring) involves complex, distinct instructions (see~\cref{fig:app:ad_like,fig:app:count,fig:app:redundancy,fig:app:stage1_llm_prompt,fig:app:stage1_vlm_prompt,fig:app:stage2,fig:app:story}), which a single model (of comparable size) struggles to handle reliably. 
Breaking the process into smaller tasks helps ensure that each instruction is properly followed.

In~\cref{tab:app:stage_ablation}, we ablate the multiple stages.
First, we evaluated a baseline where the VLM directly generates a single AD (row 4). 
This performs noticeably worse, with a StoryRecall of $2.38$ and substantially higher repetition, highlighting the limitations of relying solely on the VLM. 
To guide the VLM toward producing a single AD without overwhelming it with detailed instructions, we replaced the three-part storyboard-style VLM prompt (\cref{fig:app:stage1_vlm_prompt}) with a simplified version that retains the core AD constraints: one present-tense sentence describing the visible action or interaction, using provided character names, with no inference or emotion, and within the word limit. 
Even then, the VLM often failed to comply with these constraints.

Second, we removed the summarisation step from our pipeline (row 2) while still generating five candidates and applying the same scoring mechanism. 
This also leads to a performance drop (StoryRecall: $2.58$ vs. $2.63$) and increased partial repeats. 
Without summarisation, the candidate descriptions are less diverse and often repeat the same visual event.
The pipeline produces candidates containing actions that violate AD guidelines (such as “talking” or “conversing”) which are then discarded during scoring, reducing the pool of viable options.
To accommodate the shift from a short paragraph to a longer and structured input, we adapted the prompt by retaining only the core AD constraints from~\cref{fig:app:stage2}: up to five present-tense sentences, grounded in visible content, distinct, standalone, and within the word limit. 
This was necessary, as the original, more detailed prompt led to verbose outputs that exceeded AD timing constraints.

Finally, row 1 is our proposed multi-stage pipeline while row 3 without multiple candidate generation and scoring was included in~\cref{tab:sota}. 
We observe that our multi-stage approach outperforms all other variants.

\begin{table}[t]
\centering
\small
\begin{tabular}{lcc}
\toprule
$r$  & StoryRecall\;$\uparrow$ & Partial Repeat\%\;$\downarrow$ \\
\midrule
1 & 2.56 & (5.05, 4.71, 4.33)  \\ 
2& 2.60 & (5.10, 4.28, 4.30)  \\ 
3 (default) & 2.63 & (5.21, 4.45, 4.18)   \\ 
4 & 2.58 & (5.20, 4.44, 4.13) \\
5 & 2.61 & (5.13, 4.45, 4.10) \\
\bottomrule
\end{tabular}

\caption{
\textbf{Varying the number of prior descriptions} ($r$) during scoring.
$r=3$ (default) gives the best StoryRecall and lowest overall repetition across positions.
}
\vspace{-0.5cm}
\label{tab:app:vary_r}
\
\end{table}
\subsection{Varying context during scoring}
\label{subsec:app:context}
\cref{tab:app:vary_r} shows results for different values of $r$, the number of prior descriptions used in~\cref{subsec:app:scoring}. 
Using $r=1$ gives the lowest partial repetition at position 1 ($5.05$) but slightly higher repetition at later positions and lower StoryRecall ($2.56$).
Performance improves with $r=2$ ($2.60$), and
plateaus beyond $r=3$.
While $r=5$ yields the lowest overall repetition ($5.13$, $4.45$, $4.10$), it does not improve StoryRecall ($2.61$).
$r=3$ achieves the best balance with the highest StoryRecall ($2.63$) and low overall repetition ($5.21$, $4.45$, $4.18$).
Exact repeats are zero in all cases.

\section{ADQA Benchmark}
\label{sec:app:adqa}

In~\cref{tab:app:adqa}, we evaluate \methodname~on 
ADQA~\citep{adqa},
a multiple-choice question-answering (MCQA) benchmark designed to assess ADs across two key dimensions: Visual Appreciation (VA) and Narrative Understanding (NU).
The benchmark is motivated by the core purposes of ADs:
(i)~enabling BVI audiences to appreciate the visual elements that enrich their experience, and
(ii)~supporting narrative understanding by conveying essential visual plot points.

\vspace{3pt}\noindent\textbf{Results.}
Q2VL achieves the highest VA accuracy ratio by densely summarising everything visible in a scene. 
However, its outputs are paragraph-length and unconstrained by AD timing, making them unsuitable for real-world narration. 
In contrast, \methodname~produces single-sentence descriptions that are concise enough to fit within AD intervals, yet still retains high VA performance ($50.34$)—second only to Q2VL.
Given its real-world applicability and strong VA score, \methodname~offers a more practical solution.
Despite scoring lower on conventional metrics like CIDEr ($13.2$) and LLM-AD-Eval ($2.17$), \methodname~ outperforms all prior methods on NU with a CC score  (correct answers grounded in the provided ADs) of $64.0$ and the highest accuracy ratio of $88.0$. 
The lower NU scores of the summarised paragraphs, compared to our method, may be attributed to the LLM’s difficulty in processing large dumps of information.
This demonstrates that our training-free, sequence-level approach produces ADs that are more functionally useful—even when not favored by similarity-based metrics—highlighting the limitations of relying solely on CIDEr-style evaluations for assessing AD quality. 
Taken together, these results establish \methodname~ as the most effective method across both VA and NU.

\section{Additional Qualitative Results}
\label{sec:app:qualitative}
\begin{figure*}
\centering
\includegraphics[width=\textwidth]{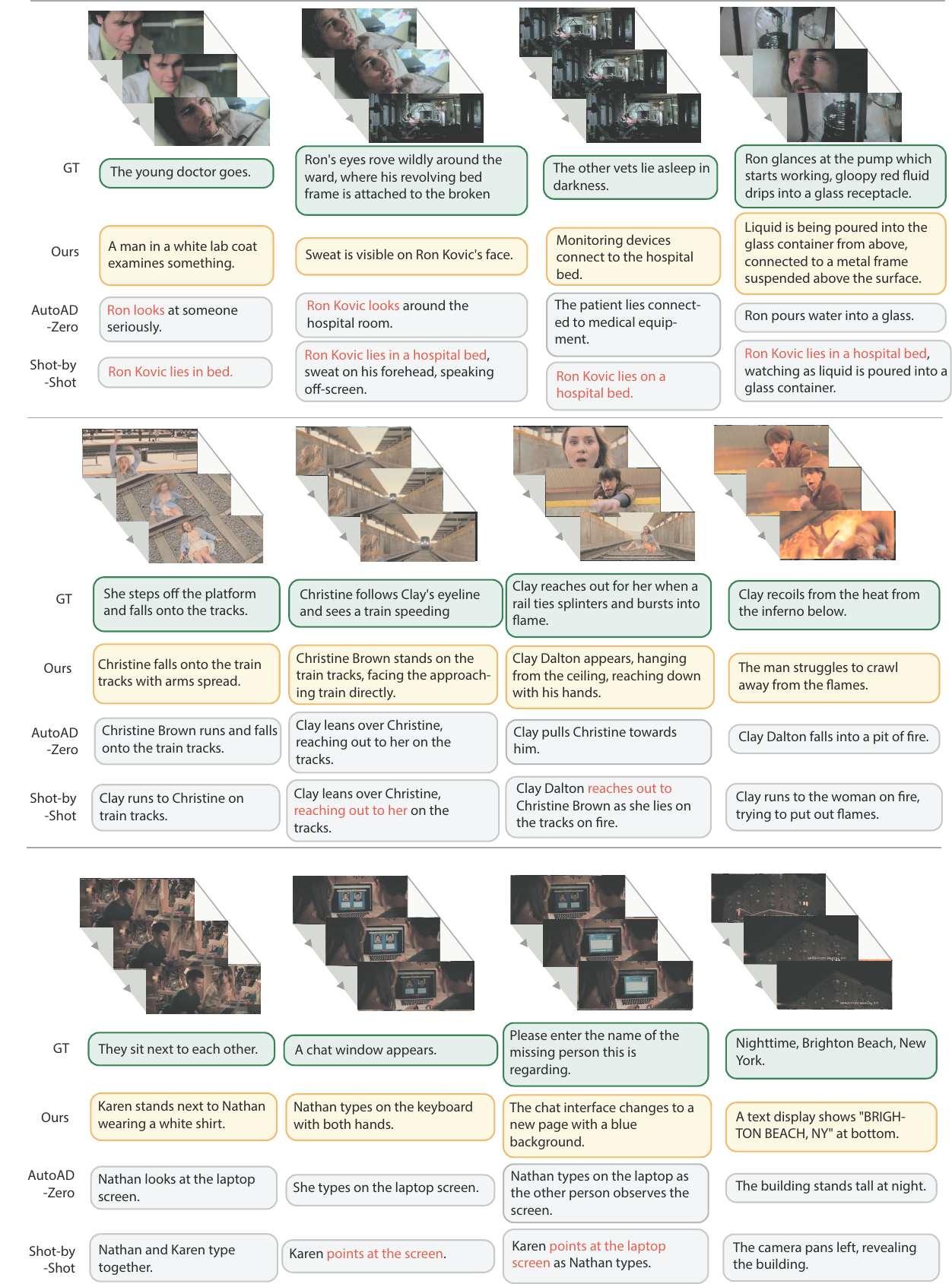}
\vspace{-0.3cm}
\caption{Qualitative comparison showing GT, our outputs, AutoAD-Zero~\cite{Xie24a} and Shot-by-Shot~\cite{shotbyshot}, with repetitions highlighted in \textcolor{BrickRed}{red}.}
\label{fig:app:qualitative}
\vspace{-0.3cm}
\end{figure*}

\begin{figure*}
\centering
\includegraphics[width=\textwidth]{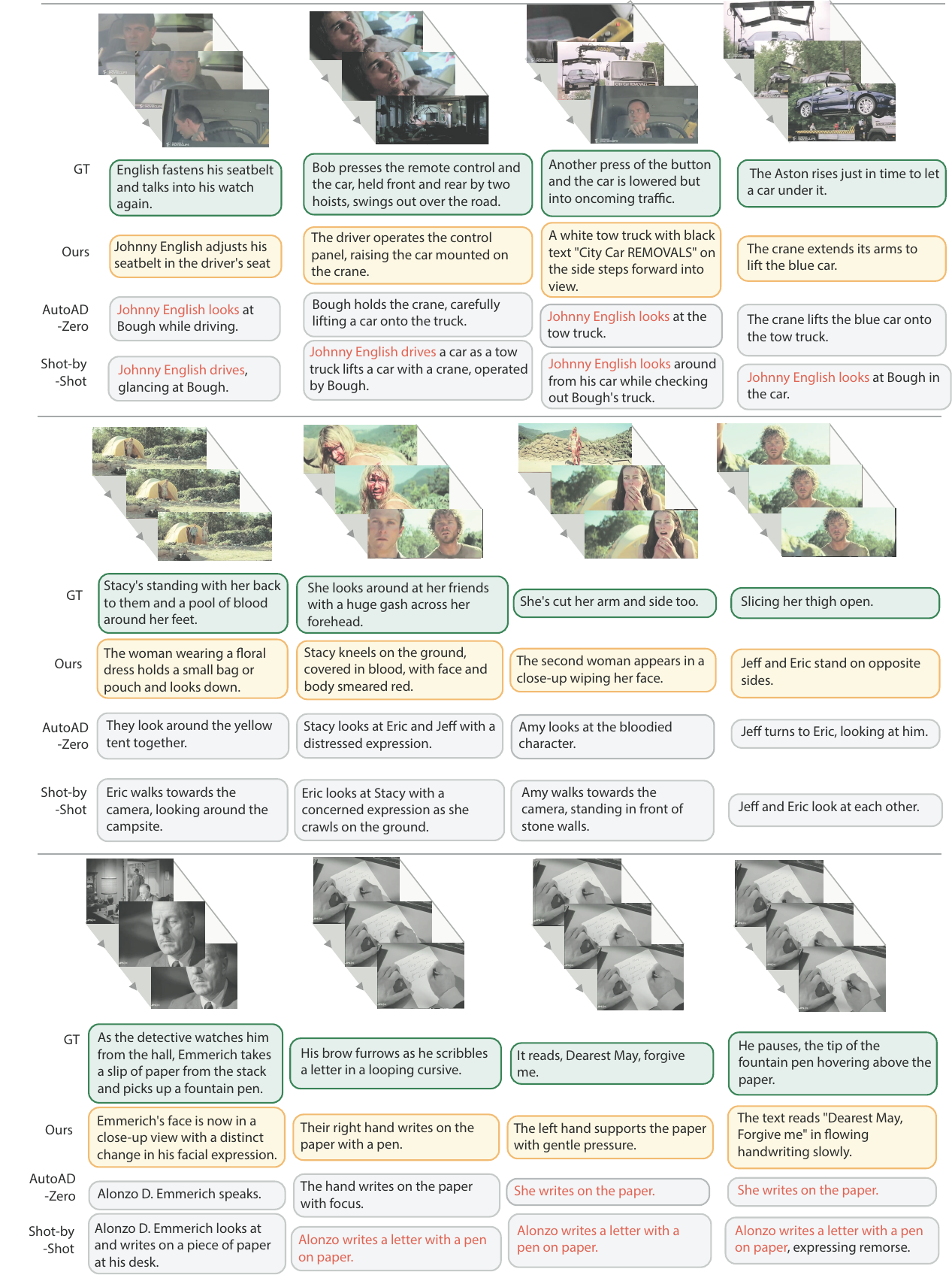}
\vspace{-0.3cm}
\caption{Qualitative comparison showing GT, our outputs, AutoAD-Zero~\cite{Xie24a} and Shot-by-Shot~\cite{shotbyshot}, with repetitions highlighted in \textcolor{BrickRed}{red}.}
\label{fig:app:qualitative2}
\vspace{-0.3cm}
\end{figure*}

\cref{fig:app:qualitative} and \cref{fig:app:qualitative2} provide additional visualisations comparing our and existing methods. 
\lstset{
  basicstyle=\footnotesize\ttfamily,  
  breaklines=true,                  
  breakatwhitespace=true,           
  columns=fullflexible,             
  keepspaces=true,                  
  showstringspaces=false,           
  frame=single,                     
  framerule=0pt,
  rulecolor=\color{black}
}

\begin{algorithm*}[htb!]
\caption{Stage I prompt for extracting structured visual descriptions.}
\label{fig:app:stage1_vlm_prompt}
\begin{lstlisting}
user_prompt = (         
    "Describe the video segment in detail using a three-part structure:\n" 
    "1. **Storyboard Description**\n"
    "   - Describe the events in the order they happen, step by step.\n" 
    "   - Use words like 'first,' 'next,' 'then,' 'finally' to show the timeline.\n"
    "   - Include clear descriptions of the location, background, objects, visible text, and any 
    changes in the setting (e.g., lights switching on, doors opening, smoke appearing).\n"
    "   - Mention any uncertainty if something is unclear.\n\n" 
    
    "2. **Character and Object Breakdown**\n"
    "{char_text}\n"
    "Break this section into these parts:\n"
    "   a. **Characters**\n"
    "      - List all visible people or animals (named or unnamed).\n"
    "      - For each character, provide two separate points:\n"
    "         i. **Visible Actions** - Big, clear movements (e.g., 'walks to the door,' 'sits on the 
    chair'). If none are seen, write 'None.'\n"
    "         ii. **Subtle Actions** - Small, visible movements or facial changes (e.g., 'raises 
    eyebrows,' 'nods'). If you believe the character seems to display an emotion (e.g., 'concerned,' 
    'nervous,' 'relieved'), clearly state the physical cues you observed that led you to that 
    interpretation. Avoid stating an emotion without citing the visible evidence. Be specific about
    gestures: say what the character does with which hand, where they point, etc. If none are seen, 
    write 'None.'\n"
    "   b. **Character-Character Interactions**\n"
    "      - List any visible interactions between characters (e.g., touching, eye contact).\n"
    "   c. **Objects**\n"
    "      - List any important objects seen in the scene.\n"
    "      - Describe how they look, where they are, and any changes they go through.\n"
    "   d. **Character-Object Interactions**\n"
    "      - Describe how characters use or touch objects (e.g., 'opens drawer,' 'picks up phone').\n"
    "   e. **Changes in Environment**\n"
    "      - List any visible changes in the surroundings (e.g., light turns off, car drives in).\n\n"

    "3. **Overall Summary**\n"
    "   - Summarize, in 1-2 sentences, the main action or event that occurs in the clip.\n"
    "   - Mention who is primarily involved (characters and/or objects).\n\n"

    "Important rules:\n"
    "- Only describe what you can see clearly.\n"
    "- Don't guess what characters are thinking or feeling unless it's visible on their face or body.\n"
    "- Say if anything is unclear or hard to see.\n"
)
\end{lstlisting}
\end{algorithm*}

\lstset{
  basicstyle=\footnotesize\ttfamily,  
  breaklines=true,                  
  breakatwhitespace=true,           
  columns=fullflexible,             
  keepspaces=true,                  
  showstringspaces=false,           
  frame=single,                     
  framerule=0pt,
  rulecolor=\color{black},
  keywordstyle=\color{blue}\bfseries,
  stringstyle=\color{teal},
  commentstyle=\color{gray},
  numberstyle=\tiny\color{gray},
  identifierstyle=\color{black},
}

\begin{algorithm*}[htb!]
\caption{Stage I prompt for narrative summarisation.}
\label{fig:app:stage1_llm_prompt}
\begin{lstlisting}
system_prompt = (
    "You are a visual summarization expert trained to convert detailed scene descriptions into 
    clean, narratable paragraphs. "
    "Your output is used as audio description for blind and visually impaired users, so it must:\n\n"
    "* Be strictly grounded in the input - only describe what is clearly visible in the source.\n"
    "* Use concrete physical verbs in present tense and active voice.\n"
    "* Avoid all emotions, sounds, dialogue, intentions, or inferences.\n"
    "* Convert static adjective phrases (like 'crossed arms') into dynamic verb phrases (like 
    'crosses arms') only if those states appear in the source.\n"
    "* Use first names **only if** explicitly and visually grounded in the input. Never guess 
    identities.\n\n"
    "Your paragraph should include all meaningful visual actions, facial or body movements, and 
    environmental changes. "
    "Keep it crisp, literal, and compliant - no more, no less."
)

user_prompt = (
    "**TASK: Convert the input scene description into one concise, literal paragraph.**\n"
    "Describe only visible, physical events - include body actions, interactions with objects, 
    posture or expression changes, and setting changes.\n\n"
    "**Guidelines**\n"
    "* Use present-tense, active voice.\n"
    "* Use concrete physical verbs (e.g., 'raises hand', 'steps forward').\n"
    "* Include subtle but visible actions (e.g., glances, nods, clenches fists).\n"
    "* Describe static elements only if they clarify the action.\n"
    "* Convert adjective states to verbs only if they appear in the input (e.g., 'crossed arms' -> 
    'crosses arms').\n"
    "* Ensure all visual events are described once and only once.\n"
    "* Use first names of the characters only if present and clearly grounded. Otherwise, use 
    specific roles or generic labels.\n"
    "* Never guess or add any action, name, or object not grounded in the source.\n"
    "**Strictly Remove**\n"
    "* Dialogue or speech verbs (speaks, talks, responds).\n"
    "* Emotions or mental states (nervous, concerned).\n"
    "* Sounds or spoken content (conversation, laughs, screams).\n"
    "* Uncertain or speculative language (e.g., 'seems to', 'possibly').\n"
    "* Camera or framing language (off-screen, towards the camera).\n"
    "* Visual markers like colored circles (red circle, green circle, blue circle, yellow circle).\n"

    "**Step 1 - Draft**\n"
    "Write a first-pass paragraph that follows all points under **Guidelines** and **Strictly 
    Remove**.\n\n"

    "**Step 2 - Refine**\n"
    "Reread your draft paragraph and revise until it passes all four checks:\n\n"
    "1. **Forbidden Terms Check**\n"
    "   * Remove any mention of emotions, thoughts, sounds, dialogue, camera angles, or visual 
    markers.\n"
    "   * Refer to the **Strictly Remove** section above.\n"
    "   * Examples to remove: 'nervous', 'appears to', 'camera pans', 'red circle', 'speaks', 
    'indicating', 'green circle'.\n\n"
    "2. **Hallucination Check**\n"
    "   * Do not add any name, object, action, or interpretation that is not grounded in the 
    input.\n"
    "   * If it's not clearly visible in the scene description, leave it out.\n\n"
    "3. **State-to-Verb Rewrite**\n"
    "   * Convert static descriptions to visible actions only when the original text implies 
    motion.\n"
    "   * Example: 'with arms crossed' becomes 'crosses arms' (if and only if supported by 
    input).\n\n"
    "4. **Coverage Check**\n"
    "   * Include every meaningful, visible action or interaction exactly once.\n"
    "   * Do not omit any relevant gestures, postures, or setting changes.\n\n"
    "Repeat this loop until the paragraph fully satisfies all checks and follows the **Guidelines** 
    and **Strictly Remove** rules.\n\n"

    "**Output**\n"
    "Output only the final paragraph, nothing else.\n\n"
    "**Scene Description**\n"
    f"{text}\n"
)
\end{lstlisting}
\end{algorithm*}

\lstset{
  basicstyle=\scriptsize\ttfamily,  
  breaklines=true,                  
  breakatwhitespace=true,           
  columns=fullflexible,             
  keepspaces=true,                  
  showstringspaces=false,           
  frame=single,                     
  framerule=0pt,
  rulecolor=\color{black},
  keywordstyle=\color{blue}\bfseries,
  stringstyle=\color{teal},
  commentstyle=\color{gray},
  numberstyle=\tiny\color{gray},
  identifierstyle=\color{black},
}

\begin{algorithm*}[htb!]
\caption{Stage II prompt for multiple candidate generation.}
\label{fig:app:stage2}
\begin{lstlisting}
system_prompt = (
    "You are a professional audio description writer.\n"
    "You convert summaries into concise, present-tense descriptions that cover only what can be 
    physically visible on screen.\n"
    "You never speculate, interpret, or describe sound or speech.\n"
    "You group related visual details into single sentences when they belong to the same moment or 
    subject.\n"
    "Your writing is literal, compact, and strictly grounded in visual facts.\n"
    "You stay within the word limit.\n"
    "Every word you use adds concrete visual value.\n"
    "You never include filler words or vague phrasing.\n"
)

user_prompt = (
    "You are given a paragraph that summarizes the visual content of a video clip. "
    "Your task is to convert this into up to 5 candidate audio descriptions (ADs) that strictly 
    follow professional AD guidelines. "
    "Each candidate must be a complete sentence in present tense, describing only what can be 
    physically visible. "
    "Do not infer or interpret anything beyond what can be explicitly seen.\n\n"
    
    "Follow these detailed instructions:\n"
    "1. Visual-Only Content:\n"
    "Describe only what can be physically visible: people, actions (both prominent and subtle), 
    interactions, objects, spatial layout, and environmental context.\n"
    "Do NOT include:\n"
    "* Emotions or internal states\n"
    "* Intentions or speculation\n"
    "* Sounds or speech-related verbs\n"
    "* Any inferred meaning or visual interpretation\n"
    "* Camera or viewer references\n"
    "* Filler words or vague phrases\n"
    "* Any colored circles (they are not meaningful scene elements)\n\n"

    "2. Sentence Structure:\n"
    "* Use present tense only.\n"
    "* Each candidate must be complete and self-contained.\n"
    "* Keep sentences concise - no filler or padding.\n"
    f"* Each sentence should aim to be exactly {num_words} words, "
    "but only include words that convey clear visual information. Do not add words just to meet the 
    target.\n\n"

    "3. Group Related Observations:\n"
    "Each sentence should describe a complete and coherent visual moment.\n"
    "Cluster visual details that naturally belong together, such as:\n"
    "* a person's action, posture, and gesture\n"
    "* a person's facial expression, gaze direction, and position in the scene\n"
    "* multiple people engaged in a single visible interaction\n"
    "* people jointly focused on a shared object or action\n"
    "* movement through a space with visible layout or surrounding elements\n"
    "* people and objects arranged in the same spatial scene\n"
    "* multiple characters described together by their clothing, positioning, or appearance\n"
    "* an object and its placement, motion, or use in the scene\n\n"
    "Avoid splitting visual details that form a single visual moment "
    "- even if they are brief or subtle. "
    "Only separate background elements if they clearly relate to different subjects or actions.\n"
    "If a gesture, facial movement, or small action is part of one visual act, group it with related 
    observations.\n\n"

    "4. Naming Conventions:\n"
    "* Use first names for characters if available.\n"
    "* Do not use full names or titles - first names are enough.\n\n"

    "5. Language Precision:\n"
    "* Do not use vague, redundant, or filler phrases such as: 'is visible', 'can be seen', 'in the 
    background'.\n"
    "* Prefer direct phrasing.\n"
    "* Avoid repetition across candidates.\n\n"

    "6. Candidate Count:\n"
    "* Generate the minimum number of candidates needed to fully cover the paragraph - up to 5.\n"
    "* If 2-4 are sufficient, stop there. Do not force 5.\n"
    f"* Remember, each candidate must aim to be {num_words} words.\n"

    "Format your output as a numbered list of 1-5 sentences, with no extra text.\n\n"

    "Here is the narrative paragraph:\n"
    f"{text} \n"

    "Now generate the candidate audio descriptions."
)    
\end{lstlisting}
\end{algorithm*}

\lstset{
  basicstyle=\footnotesize\ttfamily,  
  breaklines=true,                  
  breakatwhitespace=true,           
  columns=fullflexible,             
  keepspaces=true,                  
  showstringspaces=false,           
  frame=single,                     
  framerule=0pt,
  rulecolor=\color{black},
  keywordstyle=\color{blue}\bfseries,
  stringstyle=\color{teal},
  commentstyle=\color{gray},
  numberstyle=\tiny\color{gray},
  identifierstyle=\color{black},
}

\begin{algorithm*}[htb!]
\caption{Stage III prompt for scoring adherence to AD guidelines}
\label{fig:app:ad_like}
\begin{lstlisting}
system_prompt = (
    "You are a precise and fair rule-checker. You only lower the score when a description clearly 
    breaks one of the defined rules.\n"
    "Be especially careful to catch:\n"
    "* Any mention of inferred emotions or internal states (e.g., 'worried', 'nervous', 'concerned', 
    'frustrated')\n"
    "* Any reference to speech or dialogue (e.g., 'talks', 'speaks', 'conversation')\n"
    "* Any reference to the camera perspective or the viewer (e.g., 'off-screen', 'away from the 
    camera')\n"
    "* Any mention of coloured circles (e.g., red/green/blue/yellow circles)\n"
)


user_prompt = (
    "You are evaluating the following description for rule violations.\n\n"
    "Description:\n"
    f"{candidate}\n\n"
    "Check whether the description breaks any of the following rules:\n"
    "* Explicit mentions of emotion or internal state (e.g., 'nervous', 'worried', 'concerned', 
    'frustrated')\n"
    "* Descriptions of speech or conversation (e.g., 'talks', 'speaks', 'discusses', conversation')\n"
    "* References to the camera perspective, screen, or the viewer (e.g., 'off-screen', 'in front of 
    the camera', 'toward the camera')\n"
    "* Mentions of coloured circles (e.g., 'red circle', 'blue circle', 'green circle', 'yellow 
    circle')\n\n"
    "Scoring:\n"
    "* 3 = Fully compliant - no rule violations\n"
    "* 2 = Partially compliant - minor violation, mostly adheres to rules.\n"
    "* 1 = Non-compliant - major rule violation(s)\n\n"
    "Important: Only give a score below 3 if the description clearly breaks one of the listed 
    rules.\n\n"
    "Clarifications:\n"
    "* A minor violation means the description still conveys meaningful visual content on its own, 
    despite the violation.\n"
    "* Do not penalize vague, brief, or underspecified descriptions.\n"
    "* Facial expressions (e.g., raising eyebrows), head movements (e.g., turning), and eye movements 
    (e.g., looking around, glancing) are permitted unless they include clear emotion words (e.g., 
    'worried', 'nervous', 'concerned', 'frustrated').\n"
    "* Mouth movements (e.g., mouth opens) are fine unless they clearly imply speech.\n"
    "* Describing a camera or screen (e.g., TV, monitor) as an object is fine - only references to 
    the camera's perspective or the viewer are violations.\n"
    "* Coloured circles are always violations.\n"
    "Examples:\n"
    "Score 3: 'The man furrows his brow and picks up a gun and a camera.' (no rule violations)\n"
    "Score 2: 'The woman in a red dress walks toward the door, looking tense.' ('looking tense' 
    describes emotion/internal state - minor violation; the main action 'walking toward the door' 
    still remains meaningful on its own)\n"
    "Score 1: 'The camera zooms in on a man with a red circle.' (references to camera and coloured 
    circle - major violations)\n\n"
    "First, list any rule-relevant observations in 1-2 sentences. If a rule is broken, note whether
    the violation is minor or major. Then, assign a score from 1 to 3.\n"
    "Output your response in the following format:\n"
    "Observations: [...]\n"
    "Score: [1-3]"
)
\end{lstlisting}
\end{algorithm*}

\lstset{
  basicstyle=\footnotesize\ttfamily,  
  breaklines=true,                  
  breakatwhitespace=true,           
  columns=fullflexible,             
  keepspaces=true,                  
  showstringspaces=false,           
  frame=single,                     
  framerule=0pt,
  rulecolor=\color{black},
  keywordstyle=\color{blue}\bfseries,
  stringstyle=\color{teal},
  commentstyle=\color{gray},
  numberstyle=\tiny\color{gray},
  identifierstyle=\color{black},
}

\begin{algorithm*}[htb!]
\caption{Stage III prompt for scoring redundancy.}
\label{fig:app:redundancy}
\begin{lstlisting}
system_prompt = (
    "You are a redundancy evaluator.\n"
    "Your job is to judge how much new information a candidate description adds beyond the previous
    description(s).\n"
    "Do not go out of your way to find small or indirect overlaps "
    "- only count something as repeated if its meaning clearly matches what was already described.\n"
    "Default to a score of 3 unless the candidate clearly, without any assumptions, repeats an event 
    (action, interaction, or visible state) that was already described.\n"
    "If no new event is introduced, assign a score of 1.\n"
)
    
user_prompt = (
    "You are checking how much new information the candidate description adds to the previous 
    description(s).\n\n"
    "Compare the candidate with the previous description(s) and judge how much of the content is 
    new.\n\n"
    "Previous description(s):\n"
    f"{current_desc}\n\n"
    "Candidate description:\n"
    f"{candidate}\n\n"
    "Assign a score from 1 to 3 based on the following criteria:\n"
    "* 1 = Almost all content is already stated - no new event is introduced.\n"
    "* 2 = Some content is new - a clearly repeated event is present, but a new event is also 
    described.\n"
    "* 3 = Most of the content is new - no events are repeated.\n"
    "Clarifications:\n"
    "* People (including their appearance if unnamed), objects, or locations do not count as repeated
    content.\n"
    "* If the candidate continues a prior event, do not treat it as repetition if it adds clearly new
    and meaningful actions or visual developments.\n"
    "* Only assign a lower score if a clearly repeated event (action, interaction, or visible state)
    is present.\n"
    "* Do not go out of your way to find subtle or indirect overlaps - only count something as 
    repeated if its meaning clearly matches what was already described.\n"
    "Examples:\n"
    "* Score 3:\n"
    "    Previous: 'A man in a red shirt is picking up a gun from the table.'\n"
    "    Candidate: 'A man in a red shirt is looking at a gun and smiling.'\n"
    "    (All events, looking at the gun, smiling, are new and not mentioned before.)\n\n"
    "* Score 2:\n"
    "    Previous: 'Jim walks toward the door.'\n"
    "    Candidate: 'Jim approaches the door and opens it with his left hand.'\n"
    "    (One event, approaching the door, is similar to walking toward it, but the second event, 
    opening the door, is clearly new.)\n\n"
    "* Score 1:\n"
    "    Previous: 'The car with its headlights on drives forward through the intersection.'\n"
    "    Candidate: 'The car moves forward with its headlights on.'\n"
    "    (The same event, moving forward with headlights on, is repeated in different words. No new
    events are added.)\n\n"
    
    "Instructions:\n"
    "Write your observations in 1-2 sentences explaining how much of the candidate's content is 
    new.\n"
    "Then assign a score from 1 to 3.\n\n"
    "Important: Default to a score of 3 unless the candidate clearly, without any assumptions, 
    repeats an event already described. If no new event is introduced, assign a score of 1.\n"
    "Output format:\n"
    "Observations: [...]\n"
    "Score: [1-3]"
)
\end{lstlisting}
\end{algorithm*}

\lstset{
  basicstyle=\footnotesize\ttfamily,  
  breaklines=true,                  
  breakatwhitespace=true,           
  columns=fullflexible,             
  keepspaces=true,                  
  showstringspaces=false,           
  frame=single,                     
  framerule=0pt,
  rulecolor=\color{black},
  keywordstyle=\color{blue}\bfseries,
  stringstyle=\color{teal},
  commentstyle=\color{gray},
  numberstyle=\tiny\color{gray},
  identifierstyle=\color{black},
}

\begin{algorithm*}[htb!]
\caption{Stage III prompt for scoring story advancement.}
\label{fig:app:story}
\begin{lstlisting}
system_prompt = (
    "You are a visual narrative progression evaluator.\n"
    "Your job is to judge how much a candidate description advances the scene beyond the previous 
    description(s).\n"
    "Treat the previous description(s) as the current state of the scene.\n"
    "Focus on new actions, interactions, or changes that clearly affect what is happening in the 
    scene.\n"
    "Minor movements or posture shifts (e.g., turning, walking, looking around) should only be scored
    higher if they cause a clear shift in focus, direction, or interaction.\n"
    "Descriptions of appearance, background, or static visual elements should receive the lowest 
    score unless they visibly affect the scene.\n"
    "Base your evaluation only on what is explicitly stated. Do not infer intent, emotions, or 
    consequences that are not shown.\n"
)

user_prompt = (
    "You assess whether a candidate description advances the visual narrative beyond the previous 
    description(s).\n\n"
    "Carefully read the previous description(s) and the candidate.\n\n"
    "Previous description(s):\n"
    f"{current_desc}\n\n"
    "Candidate description:\n"
    f"{candidate}\n\n"
    "Evaluate what new visual information the candidate explicitly adds. "
    "Look for new actions, interactions, or visually meaningful changes.\n"
    "Scoring Criteria:\n"
    "* 5 = Major action, event, or change that clearly advances the scene.\n"
    "  Example: 'The man pulls the trigger, and the gun fires.' (Highly significant change in the 
    scene)\n"
    "* 4 = Clear action, interaction, or change that adds meaningful development to the scene.\n"
    "  Example: 'She picks up the phone from the table.' (Initiates a new event)\n"
    "* 3 = Minor action or movement that slightly advances the scene by shifting focus, direction, or
    interaction.\n"
    "  Example: 'The boy steps away from the table.' (shift in position)\n"
    "* 2 = Minor gestures or visual details that add tone or context but do not affect what is 
    happening in the scene.\n"
    "  Example: 'The woman sits at her desk.' (No change in the scene)\n"
    "* 1 = Static visual detail with no narrative impact.\n"
    "  Example: 'A lamp rests on the side table.' (No change in the scene)\n\n"
    "Important:\n"
    "* Score based on whether the candidate changes the current state of the scene.\n"
    "* Descriptions that visibly change the course of events or introduce new interactions should 
    score higher.\n"
    "* Minor actions or gestures with no effect on others or the unfolding situation should score 2
    or lower.\n"
    "* Purely descriptive details about appearance, background, or already-known elements should 
    score 1.\n"
    "* Onscreen text should be scored by its narrative impact. If it introduces new facts or reframes
    the scene, it may merit a 3-5.\n"
    "* Only use the information explicitly shown in the candidate. Do not assume anything beyond what
    is described.\n"
    "Describe what the candidate contributes to the ongoing scene in 1-2 sentences. Then assign a 
    score from 1 to 5.\n"
    "Output Format:\n"
    "Observation: [...]\n"
    "Score: [1-5]"
)
\end{lstlisting}
\end{algorithm*}

\lstset{
  basicstyle=\footnotesize\ttfamily,  
  breaklines=true,                  
  breakatwhitespace=true,           
  columns=fullflexible,             
  keepspaces=true,                  
  showstringspaces=false,           
  frame=single,                     
  framerule=0pt,
  rulecolor=\color{black},
  keywordstyle=\color{blue}\bfseries,
  stringstyle=\color{teal},
  commentstyle=\color{gray},
  numberstyle=\tiny\color{gray},
  identifierstyle=\color{black},
}

\begin{algorithm*}[htb!]
\caption{Stage III prompt for counting visual elements.}
\label{fig:app:count}
\begin{lstlisting}
system_prompt = (
    "You are an expert in structured scene parsing.\n"
    "Extract only explicit, observable, and non-redundant visual details from a description.\n"
    "Each item must be counted exactly once - no duplicates within or across categories.\n"
    "Only include elements that are clearly described and visually relevant to the described event.\n"
    "Participants and Actions must play a central role in the described event. "
    "If an entity or action is not clearly central, demote it to 'Other Details'.\n"
)
  
user_prompt = (
    "You are given a scene description. "
    "Extract and count only the most visually **salient** elements under the following two 
    categories:\n\n"
    
    "----\n\n"
    
    "1. **Participants**"
    "* Include only people, animals, or objects that play a **visually central and narratively 
    important role**.\n"
    "* Do **not** include someone just for being present or named. They must be doing something 
    important, or something important must be happening to them. \n"
    "* Prioritize scenes with **multiple active entities** - especially if they are interacting 
    meaningfully. \n"
    "* Ask: *Would this participant make the moment feel different if removed?*\n\n"
    
    "**Valid examples:**\n"
    "- woman covered in blood\n"
    "- man pointing a gun\n"
    "- child gripping a torn photo\n\n"
    
    "**Invalid examples (unless clearly emphasized):**\n"
    "- person walking\n"
    "- woman seated in the background\n"
    "- man standing\n\n"
    
    "----\n\n"
    
    "2. **Other Details**\n"
    "* Include only **striking descriptive elements** - things that change the tone, reveal something
    dramatic, or stand out visually.\n"
    "* Focus on things like blood, injuries, fire, smoke, damage, or strong emotional expressions.\n"
    "* Do **not** include ordinary background elements, red/green circles, or routine 
    clothing/furniture unless the sentence highlights them as important.\n"
    "* Ask: *Would a blind viewer miss something essential if this detail were skipped?*\n\n"
    
    "**Valid examples:**\n"
    "- blood on the floor\n"
    "- shattered glass underfoot\n"
    "- smoke billowing from a doorway\n\n"
    
    "**Invalid examples (unless clearly emphasized):**\n"
    "- red circle, green circle\n"
    "- lamp, couch, hat visible in the background\n\n"
     
    "----\n\n"
    
    "**Important Guidelines**\n"
    "- Leave categories empty unless something clearly stands out.\n"
    "- Count only what is **explicitly stated**, not inferred.\n"
    "- Do **not** list anything generic or background unless the sentence signals its importance.\n"
    "- Each detail must be **distinct** and appear in only one category.\n\n"
    
    "**Output Format (strict):**\n"
    "Participants: <comma-separated list> - <count>\n"
    "Other Details: <comma-separated list> - <count>\n\n"
    
    "Now extract salient visual content from the following description:\n"
    
    "Description:\n"
    f"{candidate}\n"
)
\end{lstlisting}
\end{algorithm*}

\end{document}